\documentclass[10pt,twocolumn,letterpaper]{article}

\usepackage{cvpr}              

\definecolor{cvprblue}{rgb}{0.21,0.49,0.74}
\usepackage[pagebackref,breaklinks,colorlinks,allcolors=cvprblue]{hyperref}

\usepackage[accsupp]{axessibility}  
\usepackage{graphicx}
\usepackage{amsmath}
\usepackage{amssymb}
\usepackage{mathabx}
\usepackage{booktabs}
\usepackage{float}
\usepackage{algorithm,algorithmic}
\usepackage{xcolor}
\usepackage{multirow}
\usepackage{array}
\newcolumntype{M}[1]{>{\centering\arraybackslash}m{#1}}
\newcolumntype{L}[1]{>{\flushleft\arraybackslash}m{#1}}
\usepackage{caption} 
\usepackage{bbm}
\usepackage{tabularx}

\def\paperID{11103} 
\def\confName{CVPR}
\def\confYear{2025}

\title{Enhancing Privacy-Utility Trade-offs to Mitigate Memorization in Diffusion Models}

\author{
    Chen Chen\textsuperscript{1} \quad 
    Daochang Liu\textsuperscript{2} \quad 
    Mubarak Shah\textsuperscript{3} \quad 
    Chang Xu\textsuperscript{1} \\
    \textsuperscript{1}School of Computer Science, Faculty of Engineering, The University of Sydney, Australia \\
    \textsuperscript{2}School of Physics, Mathematics and Computing, The University of Western Australia, Australia \\
    \textsuperscript{3}Center for Research in Computer Vision, University of Central Florida, USA \\
    {\tt\small \{cche0711@uni., c.xu@\}sydney.edu.au} \hspace{0.2mm}
    {\tt\small daochang.liu@uwa.edu.au} \hspace{0.2mm}
    {\tt\small shah@crcv.ucf.edu}
}

\begin{document}
\maketitle
\begin{abstract}
Text-to-image diffusion models have demonstrated remarkable capabilities in creating images highly aligned with user prompts, yet their proclivity for memorizing training set images has sparked concerns about the originality of the generated images and privacy issues, potentially leading to legal complications for both model owners and users, particularly when the memorized images contain proprietary content. Although methods to mitigate these issues have been suggested, enhancing privacy often results in a significant decrease in the utility of the outputs, as indicated by text-alignment scores. To bridge the research gap, we introduce a novel method, PRSS, which refines the classifier-free guidance approach in diffusion models by integrating prompt re-anchoring (PR) to improve privacy and incorporating semantic prompt search (SS) to enhance utility. Extensive experiments across various privacy levels demonstrate that our approach consistently improves the privacy-utility trade-off, establishing a new state-of-the-art.
\end{abstract}
\vspace{-0.7cm}
\section{Introduction}
\begin{figure}[tb]
  \centering
  \includegraphics[width=1.0\linewidth]{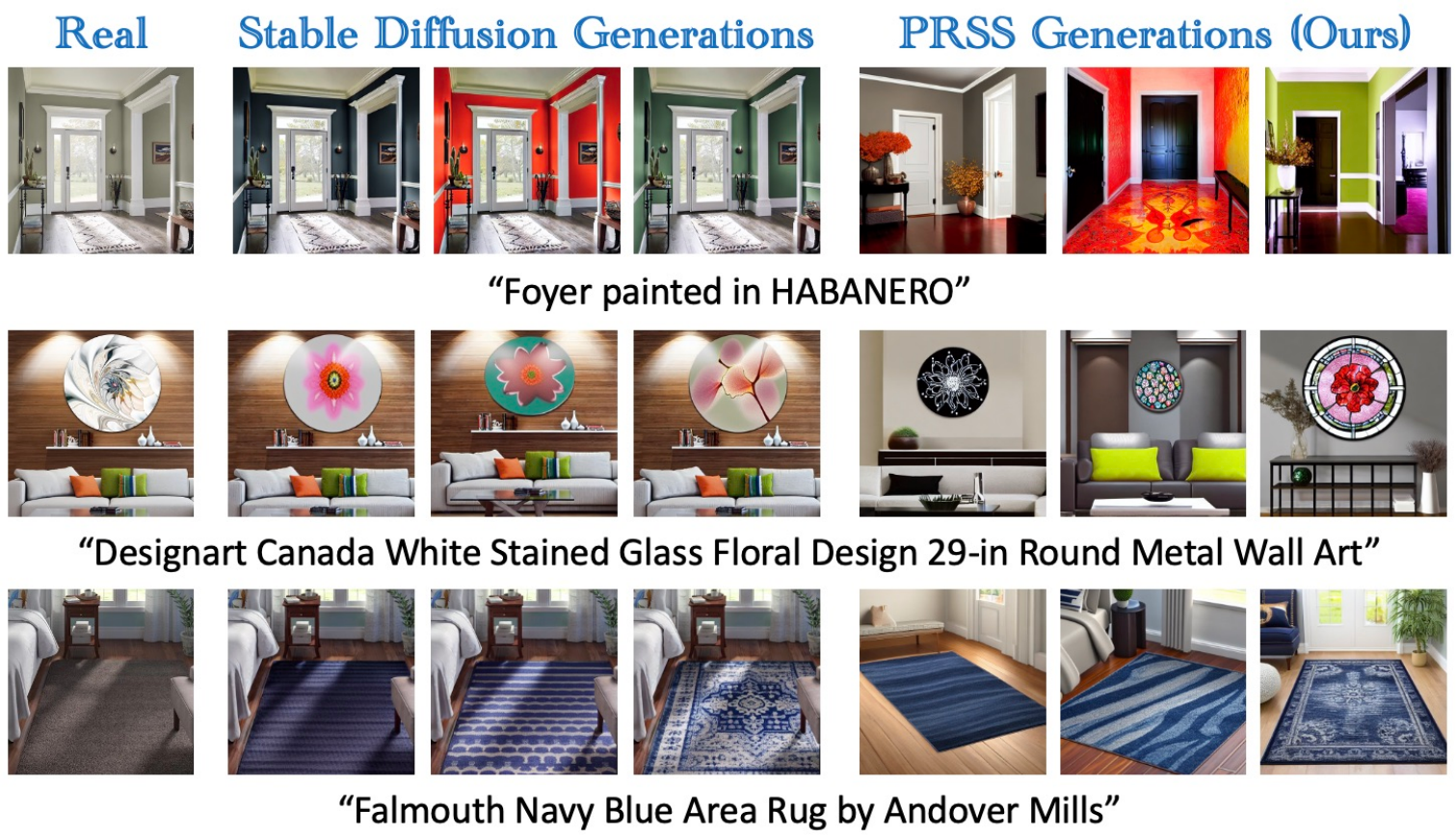}
  \caption{Examples of \textit{Local Memorization}: Stable Diffusion generations (columns 2–4) can replicate local regions of training images (column 1). Our PRSS method effectively mitigates memorization while maintaining strong alignment with the input prompts (columns 5-7).
  }
  \label{fig:teaser_localmem}
\vspace{-0.4cm}
\end{figure}

\begin{figure}[tb]
  \centering
  \includegraphics[width=1.0\linewidth]{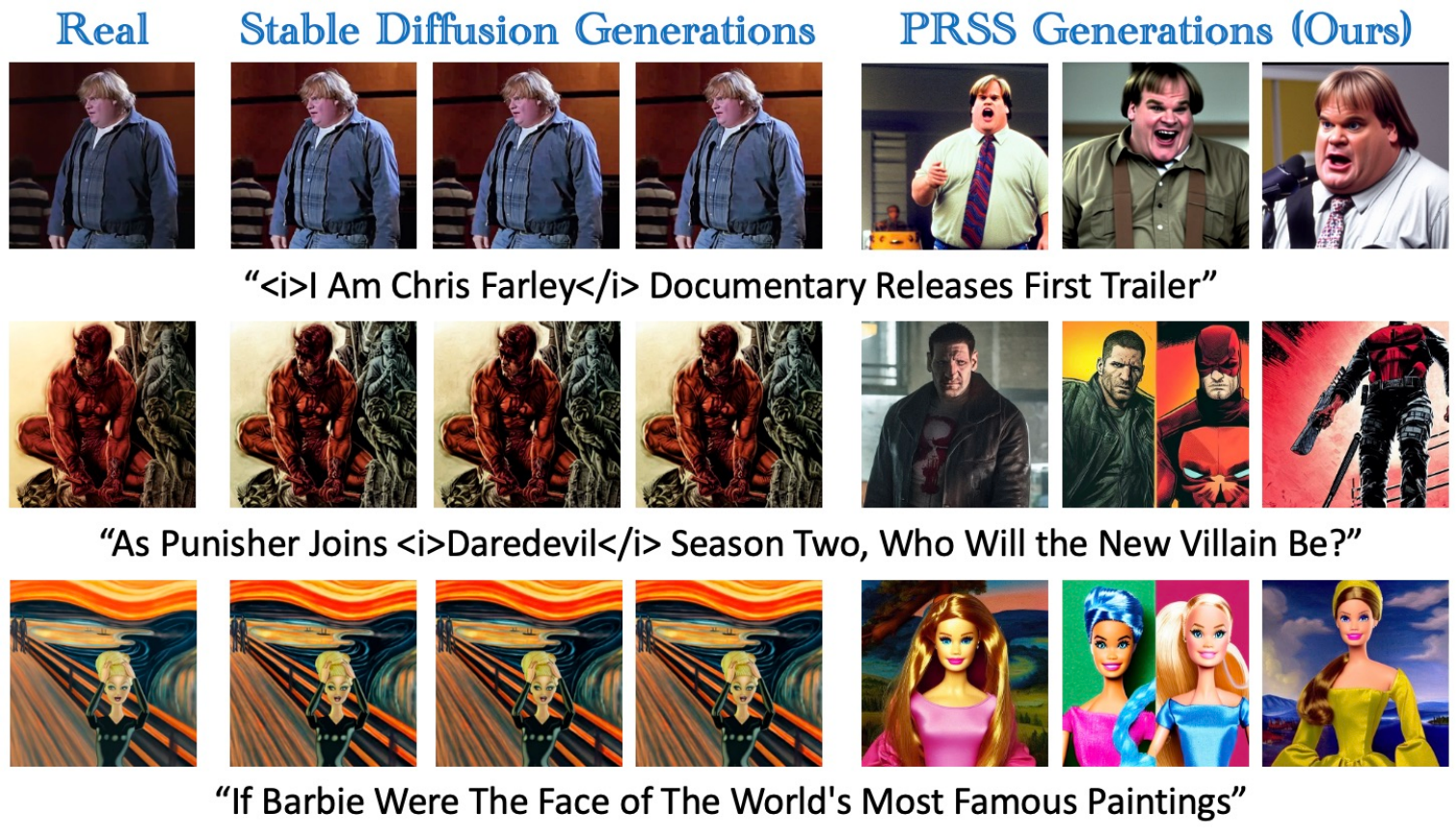}
  \caption{Examples of \textit{Global Memorization}: Stable Diffusion generations (columns 2–4) can replicate the entire training images (column 1). Our PRSS method effectively mitigates memorization while maintaining strong alignment with the input prompts (columns 5-7).
  }
  \label{fig:teaser_globalmem}
\vspace{-0.6cm}
\end{figure}

\vspace{-0.1cm}
\label{sec:intro}
Leveraging classifier-free guidance (CFG)~\cite{cfg}, text-to-image diffusion models like Stable Diffusion~\cite{sd_2022_cvpr} and Midjourney~\cite{midjourney} are now capable of generating highly realistic images that closely align with user-provided text prompts. This capability has propelled their popularity, leading to widespread use and distribution of their generated images.
However, recent research~\cite{carlini_2023_usenix, somepalli_2023_cvpr, Webster2023, chen_be} has revealed a critical issue: these models can memorize training data, leading them to reproduce parts of images, such as foregrounds or backgrounds (local memorization, see \cref{fig:teaser_localmem}), or even entire images (global memorization, see \cref{fig:teaser_globalmem}) during inference, instead of generating genuinely novel content.
When the training data includes sensitive or copyrighted material, these memorization issues can infringe on copyright laws without notifying either the model's owners or users or the copyright holders of the replicated content. The risk is especially substantial given the extensive use of these models and their reliance on massive datasets, such as LAION~\cite{LAION5B}, which contain billions of web-scale images and are impractical to thoroughly review or filter manually.
This risk is further highlighted by real-world cases, where several artists have filed lawsuits, arguing that models like Stable Diffusion act as ``21st-century collage tools” that remix their copyrighted works, implicating Stability AI, DeviantArt, Midjourney, and Runway AI. Recognizing the potential for unauthorized reproductions, Midjourney has even banned prompts containing the term ``Afghan" to prevent the generation of the copyrighted Afghan Girl photograph. Yet, as \cite{wen_afghan} demonstrates, such restrictions alone are insufficient to fully prevent the reproduction of copyrighted images. This underscores the urgent need for timely effective mitigation strategies to address these concerns.
\vspace{-0.2cm}

In response to these legal challenges, recent initiatives \cite{somepalli_2023_neurips, wen_2024_iclr, chen-amg, ren2024unveiling, chen_be} have focused on developing strategies to minimize memorization, achieving notable success. These approaches vary in scope, with some targeting the model's training phase and others making adjustments during inference.
While training-phase strategies are theoretically viable, they lack practical significance as they predominantly experiment on a small 10k subset of the entire LAION5B dataset due to the impracticality of fine-tuning on the full dataset — a constraint stemming from computational limitations. 
In contrast, inference-phase strategies, which involve modifications only during inference, offer computational efficiency and adaptability, as they require no additional training or fine-tuning. This makes them highly applicable across pre-trained diffusion models, regardless of specific training configurations.
Despite their promise, inference-time strategies can also suffer from computational constraints. For example, the approach proposed by~\cite{chen-amg} involves searching the entire LAION dataset for the nearest neighbor to optimize the generated latent representation during each denoising step.
Other inference-time strategies~\cite{somepalli_2023_neurips, wen_2024_iclr, ren2024unveiling, chen_be} that avoid nearest neighbor search demonstrate high efficiency but only moderate effectiveness, leaving a substantial gap in optimizing the privacy-utility trade-off. 
In particular, these methods often require significant compromises in text-image alignment to reduce memorization and conversely, struggle to sufficiently suppress memorization when trying to preserve alignment.


To address these limitations, we carefully analyze why existing strategies struggle to achieve an optimal privacy-utility trade-off, uncovering critical insights that the two inputs (text-conditional and unconditional predictions) used in the CFG framework are both suboptimal for mitigating memorization, even though they are standard for most other tasks in Stable Diffusion models. Specifically, we find that the text-conditional term in CFG, when optimized for privacy by prompt engineering, can significantly reduce text alignment, indicating the need for alternative methods to find privacy-preserving yet intent-aligned text-conditioning. Additionally, we observe that the unconditional term in CFG is less effective as an anchor for privacy preservation, suggesting that alternative anchoring points that enhance privacy are necessary.
Building on these insights, we introduce \textit{PRSS}, a novel method that incorporates two complementary strategies to refine the widely used CFG approach for improved memorization mitigation. Our method includes 
(1) \textit{prompt re-anchoring}, designed to preserve high privacy with a smaller utility cost, which re-anchors the unconditional component in CFG using the memorized prompt to consistently guide generations away from memorization during the inference process;
(2) \textit{semantic prompt search}, designed to preserve high utility with minimal privacy compromise, which optimizes the prompt away from the memorized one in the language space while keeping similar semantics.
Our experimental results reveal that employing both strategies synergistically can consistently outperform baseline methods in achieving superior privacy-utility trade-offs at different privacy levels. In particular, this results in higher text alignment (utility) and substantially fewer memorized generations (improved privacy). Furthermore, ablation studies validate the efficacy of each component within our strategy.
From an implementation standpoint, our strategies are simple, efficient, and effective, necessitating only adjustments to the existing CFG equation for updating the noise prediction during inference. This integration is seamless, avoiding disruption to the inference workflow and eliminating the need for additional re-training, fine-tuning, or searching over the training data.
\section{Preliminaries}
\label{sec:preliminaries}
\subsection{Classifier-Free Guidance in Diffusion Models}
\label{sec:guidance}
Diffusion models \cite{ddpm, iddpm_2021_icml} are popular generative models that consist of a forward diffusion process that consistently adds Gaussian noise to real images $x_0$ for $T$ steps based on the following conditional function, resulting in $x_T \sim \mathcal{N}(0, \mathbf{I})$:
\begin{equation}
q(x_t|x_{t-1}) = \mathcal{N}(x_t; \sqrt{1 - \beta_{t}} x_{t-1}, \beta_t \mathbf{I})
  \label{eq:pre1},
\end{equation}
where $x_t$ represents the noised version of $x_0$ after adding $t$ noise, and $\{\beta_t\}_{t=1}^{T}$ is the schedule that controls the magnitude of added noise at time $t$.

This is followed by the reverse process, in which a denoiser model is trained to predict the distribution (i.e., mean $\mu_{\theta}(x_t)$ and variance $\sigma_{\theta}^2(x_t)\mathbf{I}$) of a cleaner version of $x_t$, denoted as $x_{t-1}$, at any given timestep $t$:
\begin{equation}
p_{\theta}(x_{t-1}|x_t) = \mathcal{N}(x_{t-1}; \mu_{\theta}(x_t), \sigma_{\theta}^2(x_t)\mathbf{I}).
  \label{eq:pre4}
\end{equation}
During training, the objective function is equivalent to have the denoiser $\epsilon_{\theta} (x_t, y, t)$ predict the noise $\epsilon_t$ to be removed, rather than the image $x_{t-1}$ at timestep $t$, conditioned on $y$, which may represent a user-provided text prompt:
\begin{equation}
\mathcal{L} = \mathbb{E}_{t \in [1,T],\epsilon \sim \mathcal{N}(0, \mathbf{I})} [\left\| \epsilon_t - \epsilon_{\theta} (x_t, y, t) \right\|_2^2].
  \label{eq:pre6}
\end{equation}

Classifier-free guidance (CFG)~\cite{cfg} is a widely used technique for balancing fidelity and diversity in the generated outputs of diffusion models, and it can be easily applied during inference. CFG operates by linearly guiding the noise prediction from the unconditional prediction, $\epsilon_{\theta} (x_t, t)$ towards the conditional prediction, $\epsilon_{\theta}(x_t, y, t)$, with a scaling factor $s > 1$ to control the degree of guidance. For simplicity, removing the timestep $t$ from the notation, the adjusted noise prediction can be defined as:
\begin{equation}
\hat{\epsilon} \leftarrow \epsilon_{\theta}(x_t) + s \cdot (\epsilon_{\theta}(x_t, y)-\epsilon_{\theta}(x_t)).
  \label{eq:pre12}
\end{equation}

\subsection{Memorization in Diffusion Models}
\label{sec:memorization}
\vspace{-0.2cm}

\textbf{Definition and Metrics.}
A widely adopted practical definition of memorization refers to generated images exhibiting high object-level similarity to training data~\cite{somepalli_2023_cvpr}. This includes the concept of reconstructive memory in diffusion models, where foreground and background objects from training images are combined, potentially with modifications such as shifting, scaling, or cropping. These reconstructions may not match any training image pixel-for-pixel but still closely resemble specific training images at the object level.
The standard metric for evaluating memorization~\cite{somepalli_2023_cvpr, somepalli_2023_neurips, wen_2024_iclr, chen-amg, ren2024unveiling, chen_be} is designed base on a pre-trained Self-Supervised Copy Detection (SSCD) model~\cite{sscd}:
\vspace{-0.1cm}
\begin{equation}
\text{SSCD}(\hat{x}) = \max_{x_i \in \mathcal{D}} \, \text{Cosine}(\mathbf{E}_{SSCD}(\hat{x}), \mathbf{E}_{SSCD}(x_i)),
\label{eq:sscd}
\vspace{-0.2cm}
\end{equation}
where for each generated image $\hat{x}$, cosine similarity is computed between its SSCD embedding $\mathbf{E}_{SSCD}(\hat{x})$ and the embeddings $\mathbf{E}_{SSCD}(x_i)$ of all training images $x_i$ in the dataset $\mathcal{D}$. The maximum similarity serves as the final SSCD score.

Most recently, \cite{chen_be} proposed a method to localize memorized regions in generated images, arguing that only memorized regions pose privacy risks, making local memorization a more meaningful concept. They introduced an efficient approach to synthesize a mask that highlights these regions without requiring access to training data. Using this mask, they proposed a new localized metric based on both object-level SSCD and pixel-level Euclidean distance:
\begin{equation}
LS(\hat{x}, x) = -\mathbbm{1}_{\text{SSCD} > 0.5} \cdot \left\| (\hat{x} - x) \circ \mathbf{m} \right\|_2,
\label{eq:ls}
\end{equation}
where $\mathbf{m}$ is the local memorization mask and $\mathbbm{1}_{\text{SSCD} > 0.5}$ is an indicator function that signifies whether the SSCD score exceeds the 0.5 threshold, which is commonly used to distinguish memorized from non-memorized content.

\textbf{Memorization Detection.}
A recent work~\cite{wen_2024_iclr} introduces an innovative approach for detecting potential memorized generations during inference by leveraging the magnitude of the difference between text-conditional $\epsilon_\theta(x_t, e_p)$ and unconditional $\epsilon_\theta(x_t, e_\phi)$ noise predictions:
\begin{equation}
m_t=\|\epsilon_\theta(x_t, e_p)-\epsilon_\theta(x_t, e_\phi)\|_2,
  \label{eq:magitude}
\vspace{-0.1cm}
\end{equation}
where memorized generations consistently exhibit larger magnitudes than non-memorized ones, making this magnitude a strong signal for detection. Furthermore, accurate detection can be achieved efficiently by evaluating the magnitude $m_{T-1}$ at the first denoising step (timestep $T-1$, highlighting the method's computational efficiency.

The contemporary work \cite{chen_be} extends this detection method by incorporating their proposed local memorization mask $\mathbf{m}$ into \cref{eq:magitude} to compute a masked magnitude, achieving state-of-the-art performance:
\begin{equation}
m'_{t} = \left\| \left( \varepsilon_\theta(x_t, e_p) - \varepsilon_\theta(x_t, e_\phi) \right) \circ \mathbf{m} \right\|_2 \bigg/ \left( \frac{1}{N} \sum_{i=1}^{N} m_i \right),
\label{eq:masked_magnitude}
\end{equation}
where $\mathbf{m}$ is the local memorization mask and $\frac{1}{N} \sum_{i=1}^{N} m_i$ is used to normalize the result by the mean of $\mathbf{m}$.
\section{Method}
\label{sec:method}

We begin with a detailed analysis in \cref{sec:tradeoff_analysis} to uncover the key challenges that hinder existing strategies from achieving an optimal privacy-utility trade-off.
Building on these insights, we evolve the baseline step by step with our proposed prompt re-anchoring (PR) strategy in Sec~\ref{sec:np}, which focuses on improving privacy, and semantic prompt search (SS) in Sec~\ref{sec:semantic}, which enhances utility. Finally, in \cref{sec:synergy}, we analyze how the combination of PR and SS achieves a synergistic effect, offering the best privacy-utility trade-off.

\subsection{Trade-off Analysis of Existing Strategies}
\label{sec:tradeoff_analysis}

\begin{figure}[tb]
  \centering
  \includegraphics[width=1.0\linewidth]{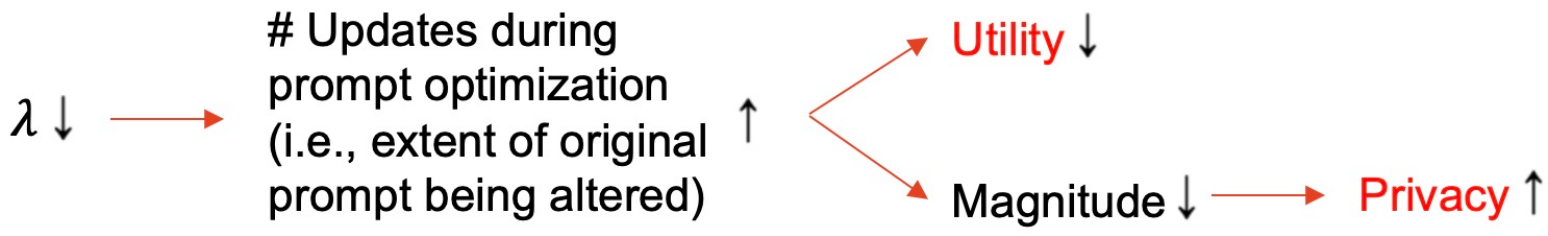}
  \caption{Illustration of the logic behind how $\lambda$ in \cref{eq:baseline} governs the privacy-utility trade-off. Currently, the privacy-utility trade-off is governed solely by how extensively the prompt is modified.
  }
  \label{fig:logic}
\vspace{-0.7cm}
\end{figure}

\begin{figure*}[t]
  \centering
  \includegraphics[width=\textwidth]{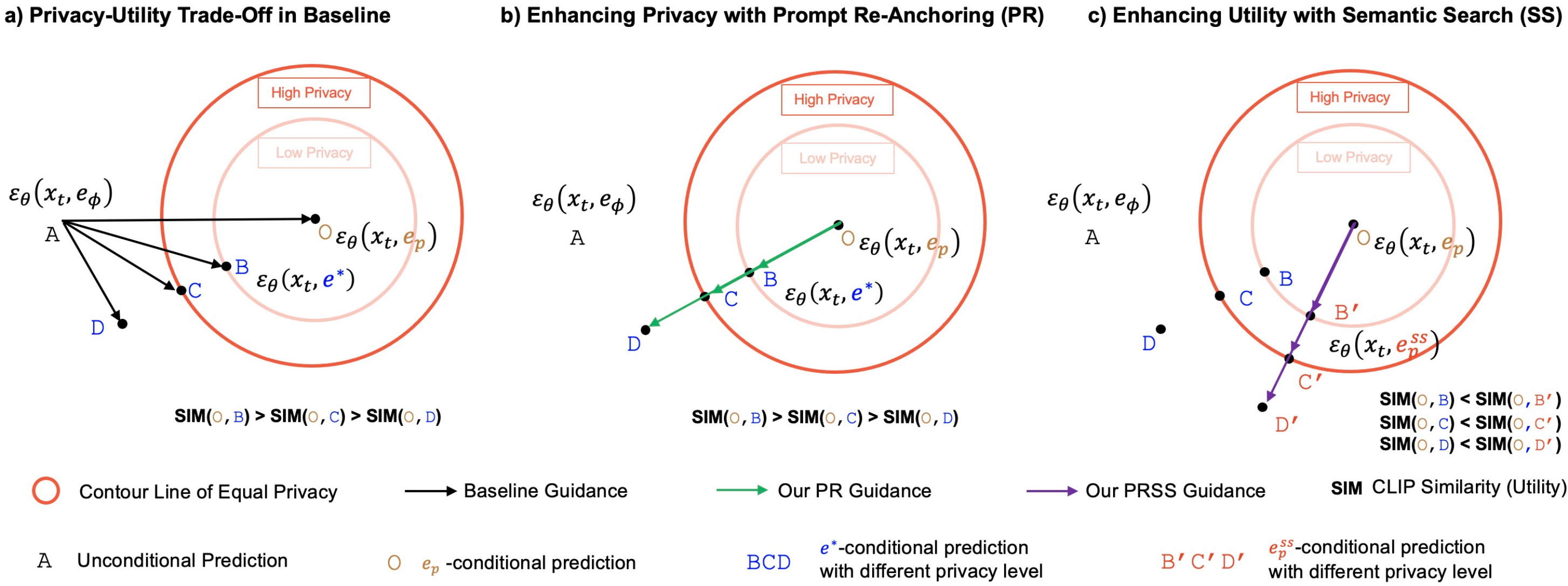}
  \vspace{-0.6cm}
  \caption{
  Illustration of our proposed strategy (PRSS) compared to the baseline prompt engineering (PE) method. 
  \textbf{a)} The privacy-utility trade-off in the baseline method, where the privacy increases at the cost of utility from $\textcolor{blue}{\texttt{B}}$ to $\textcolor{blue}{\texttt{C}}$ to $\textcolor{blue}{\texttt{D}}$.
  \textbf{b)} Our proposed PR guidance, i.e, the green arrow, drives the image out of the low-privacy region more efficiently than the baseline guidance, i.e, the black arrows. 
  \textbf{c)} Our proposed SS can enhance utility, by finding points $\textcolor{orange}{\texttt{B'}\texttt{C'}\texttt{D'}}$ that are at the same level of privacy as $\textcolor{blue}{\texttt{B}\texttt{C}\texttt{D}}$ but with better utility.
  In summary, using the proposed PR and SS jointly can enhance the privacy-utility trade-off to mitigate memorization in diffusion models.
}
  \label{fig:method}
\vspace{-0.5cm}
\end{figure*}

Our method is built upon the widely-used classifier-free guidance (CFG) by treating the input text prompt from user $\textcolor{brown}{e_p}$ as the condition $y$ in \cref{eq:pre12} and rewriting the unconditional noise as a null condition $e_\phi$,
\begin{equation}
\hat{\epsilon} \leftarrow {\epsilon}_{\theta}(x_t, e_{\phi}) + s ({\epsilon}_{\theta}(x_t, \textcolor{brown}{e_p}) - {\epsilon}_{\theta}(x_t, e_{\phi})).
  \label{eq:baseline-1}
\end{equation}
The one-step prompt engineering (PE) based memorization mitigation method proposed by \cite{wen_2024_iclr} then counteract memorization by reformulating \cref{eq:baseline-1} as:
\begin{equation}
\begin{split}
\hat{\epsilon} \leftarrow [{\epsilon}_{\theta}(x_t, e_{\phi}) + s ({\epsilon}_{\theta}(x_t, \textcolor{brown}{e_p}) - {\epsilon}_{\theta}(x_t, e_{\phi}))] \mathbbm{1}_{\{m_{T-1} < \lambda\}} \\
+ [{\epsilon}_{\theta}(x_t, e_{\phi}) + s ({\epsilon}_{\theta}(x_t, \textcolor{blue}{e^*}) - {\epsilon}_{\theta}(x_t, e_{\phi}))] \mathbbm{1}_{\{m_{T-1} > \lambda\}},
  \label{eq:baseline}
\end{split}
\end{equation}
where $m_{T-1}=\|\epsilon_\theta(x_{T-1}, e_p)-\epsilon_\theta(x_{T-1}, e_\phi)\|_2$ represents the magnitude at the initial step $T-1$ as per \cref{eq:magitude}, $\epsilon_\theta(x_t, \textcolor{brown}{e_p})$ and $\epsilon_\theta(x_t, \textcolor{blue}{e^*})$ are the noise predictions conditioned on the user prompt embedding $\textcolor{brown}{e_p}$ and engineered prompt embedding \textcolor{blue}{$e^*$} (optimized based on the gradient computed from the first-step magnitude $m_{T-1}$), and $\epsilon_\theta(x_t, e_\phi)$ is the unconditional noise prediction.
This strategy detects potential memorization for a given input prompt based on the magnitude signal $m_{T-1}$ with a threshold $\lambda$, and undertakes soft prompt engineering (\ie, optimizes the prompt embeddings) if potential memorization is detected.
In this context, if $m_{T-1}$ is under $\lambda$, the prompt is considered safe from memorization, and the generation proceeds with the original user prompt \textcolor{brown}{$e_p$}, mirroring Stable Diffusion's output.
Conversely, if $m_{T-1}$ is above $\lambda$, the prompt is then identified as likely to produce memorized images, thus \textcolor{brown}{$e_p$} is optimized to and replaced with an engineered prompt \textcolor{blue}{$e^*$} in the conditional prediction while keeping the unconditional prediction $\epsilon_\theta(x_t, e_\phi)$ as is following the classical CFG equation.
Specifically, \textcolor{blue}{$e^*$} is optimized to reduce the first-step magnitude $m_{T-1}$ using gradient descent and the optimization halts once the magnitude $m_{T-1}$ falls below $\lambda$, marking the engineered prompt as less prone to memorization.
Both detection and prompt optimization occur only during the initial inference step $T-1$, rather than throughout the entire inference process, enhancing efficiency.

\textbf{Privacy-utility trade-off}. 
\cref{eq:baseline} suffers from a severe privacy-utility trade-off as in \cref{fig:logic}. 
Opting for a lower $\lambda$ improves the privacy (risk of memorization) of outputs at the expense of utility (text-alignment). 
It requires more substantial optimization of the input prompt, aiming to achieve a magnitude $m_{T-1}$ lower than $\lambda$, thereby enhancing privacy. However, such extensive modifications to the prompt may lead to a loss of its original meaning, compromising utility significantly.
Conversely, a higher $\lambda$ upholds the utility of generated outputs but at the cost of privacy. 

\textbf{Geometric analysis.}
\cref{fig:method} (a) visualizes the privacy-utility trade-off in \cref{eq:baseline} as a function of $\lambda$, where varying levels of $\lambda$ define distinct contours corresponding to target magnitudes and privacy levels during prompt engineering (PE). Outer contours represent greater prompt modifications, which enhance privacy.
This optimization shifts the prompt embedding from the original memorized prompt, \textcolor{brown}{$e_p$}, located at point \texttt{O}, to the engineered prompt, \textcolor{blue}{$e^*$}, at points \texttt{B}, \texttt{C}, or \texttt{D}. The number of PE iterations determines the distance of the prompt from point \texttt{O}: more iterations lead to greater divergence, achieving higher privacy levels. The process halts once the desired privacy level is attained.

\textbf{Overview of our improvements}.
Our trade-off analysis reveals that existing strategies rely solely on prompt modification methods such as prompt engineering to enhance privacy, which is suboptimal and significantly compromises utility. 
This highlights an urgent need to propose alternative methods to improve privacy and preserve utility.
To address this, we introduce two meticulously crafted mitigation strategies that only require refining the guidance components in CFG to achieve a superior privacy-utility trade-off: (1) \textit{prompt re-anchoring (PR)} and (2) \textit{semantic prompt search (SS)}. 
Specifically, SS enhances utility with minimal privacy cost, while PR serves as an effective alternative to improve privacy, which has the potential of costing less utility when used in combination with SS.

\vspace{-0.1cm}
\subsection{Enhancing Privacy Through Prompt Re-Anchoring}
\label{sec:np}
\vspace{-0.1cm}

One of our objectives is to safeguard privacy while preserving as much utility as possible.
Recall that in \cref{eq:baseline} enhancing privacy requires reducing the magnitude, which correlates with a significant decrease in utility. 
It is important to note that the utility loss stems not from the reduction in magnitude itself, but from the baseline’s inefficient approach to achieving that reduction.
Visually, in \cref{fig:method}, predictions that lie on the same magnitude contour may exhibit notably different levels of utility.
This observation motivates us to ponder: Is it possible to reduce magnitude through a more efficient approach (incurring a lower utility cost) compared to the baseline method of prompt engineering?
This insight has spurred the creation of our prompt re-anchoring (PR) strategy, which aims to reduce magnitude by steering predictions away from memorized images.

We argue that \textcolor{brown}{$e_p$} is discarded too hastily in \cref{eq:baseline} when the memorization is detected ($m_{T-1} > \lambda$). 
Instead, recognizing \textcolor{brown}{$e_p$} as a valuable anchor point — its conditional generation can trace the memorization path — we use \textcolor{brown}{$e_p$} to re-anchor the generation process, by replacing the unconditional prediction in CFG with the \textcolor{brown}{$e_p$}-conditioned prediction to divert the generation away from memorization.
Thereby, \cref{eq:baseline} is re-written to the following form,
\begin{equation}
\begin{split}
\hat{\epsilon} \leftarrow [{\epsilon}_{\theta}(x_t, e_{\phi}) + s ({\epsilon}_{\theta}(x_t, \textcolor{brown}{e_p}) - {\epsilon}_{\theta}(x_t, e_{\phi}))] \mathbbm{1}_{\{m_{T-1} < \lambda\}} \\
+ [{\epsilon}_{\theta}(x_t, \textcolor{brown}{e_p}) + s ({\epsilon}_{\theta}(x_t, \textcolor{blue}{e^*}) - {\epsilon}_{\theta}(x_t, \textcolor{brown}{e_p}))] \mathbbm{1}_{\{m_{T-1} > \lambda\}}.
  \label{eq:np}
\end{split}
\end{equation}

\textbf{Analysis: Enhancing Privacy with Less Utility Cost.}
The baseline approach in \cref{eq:baseline} to enhancing privacy necessitates significant utility sacrifices, primarily due to the need for more extensive gradient updates during prompt optimization to obtain \textcolor{blue}{$e^*$}. This leads to substantial modifications of the original user prompt and a large gap between \textcolor{blue}{$e^*$} and \textcolor{brown}{$e_p$}, causing the prompt to diverge significantly from the user's initial intent and thus undermine the text alignment.
In contrast, our prompt re-anchoring strategy employs the intuition from the original CFG approach to guide generations toward a preferred output while distancing them from an undesirable one, effectively achieved through linear interpolation between two generations. Uniquely, our strategy designates the undesired generation as the memorized one that enforces conditioning on the original user prompt \textcolor{brown}{$e_p$}, rather than relying on unconditional conditioning $e_{\phi}$. 
This approach steers generations away from memorized content more efficiently with the contrastive direction, which therefore requires fewer gradient updates to obtain \textcolor{blue}{$e^*$} and thus less alternation on the prompt for the same privacy enhancement level.
Our previous question is hence answered.

\cref{fig:method} represents a geometric interpretation of our prompt re-anchoring strategy. Classical CFG directs $\hat{\epsilon}$ away from unconditional (point \texttt{A}) towards conditional predictions (\texttt{O} for \textcolor{brown}{$e_p$}, and \texttt{B}, \texttt{C}, \texttt{D} for \textcolor{blue}{$e^*$}) as in the left figure. 
PR, however, guides $\hat{\epsilon}$ from the memorized prediction (\texttt{O}) towards less or unmemorized predictions with lower magnitude (\texttt{B}, \texttt{C}, \texttt{D}), enhancing privacy, as in the middle figure.

\textbf{Analysis: Continuous Memorization Diversion without Loss of Efficiency}.
Although \cite{wen_2024_iclr} found that minimizing the initial magnitude $m_{T-1}$ can indirectly reduce magnitudes in later steps, we observe that the magnitude can still spike and memorization may still occur later in the inference process.
For example, as illustrated in \cref{fig:analysis_pr_synergy} (in Supplementary Material), while the engineered prompt achieves a low magnitude initially, it fails to ensure that this low level is maintained throughout subsequent inference steps, potentially leading back to high magnitudes indicative of Stable Diffusion's memorized outputs.
To counteract this, the PR strategy can continuously divert generations away from initially identified memorized conditions (point \texttt{O}) across the entirety of the inference process as shown in \cref{fig:analysis_pr_synergy} (in Supplementary Material). 
The effectiveness is also evident from the qualitative appearance of the final generated image, and its quantitative similarity score as in \cref{fig:analysis_pr_synergy} (in Supplementary Material).

\subsection{Enhancing Utility Through Semantic Prompt Search}
\label{sec:semantic}
In this section, we aim to propose a strategy to enhance utility with minimal privacy cost.
Recall that low utility is primarily due to the increased difference between the user prompt and the engineered prompt.
We suggest conducting semantic search to find a prompt \textcolor{red}{$e_{p}^{ss}$} semantically similar to the original user prompt \textcolor{brown}{$e_p$} but with a lower risk of memorization, aiming to replace the engineered prompt \textcolor{blue}{$e^*$} that results in the unsatisfactory trade-off.
\cref{eq:np} is thus further re-formulated to the following form after incorporating semantic prompt search strategy in addition to the prompt re-anchoring strategy,
\begin{equation}
\begin{split}
\hat{\epsilon} \leftarrow [{\epsilon}_{\theta}(x_t, e_{\phi}) + s ({\epsilon}_{\theta}(x_t, \textcolor{brown}{e_p}) - {\epsilon}_{\theta}(x_t, e_{\phi}))] \mathbbm{1}_{\{m_{T-1} < \lambda\}} \\
+ [{\epsilon}_{\theta}(x_t, \textcolor{brown}{e_p}) + s ({\epsilon}_{\theta}(x_t, \textcolor{red}{e_{p}^{ss}}) - {\epsilon}_{\theta}(x_t, \textcolor{brown}{e_p}))] \mathbbm{1}_{\{m_{T-1} > \lambda\}}.
  \label{eq:gpt}
\end{split}
\end{equation}
In \cref{eq:gpt}, the engineered prompt \textcolor{blue}{$e^*$} from \cref{eq:baseline} and \cref{eq:np} is replaced by \textcolor{red}{$e_{p}^{ss}$}.
The search is conducted in the linguistic space to inherently preserve semantics.
We use a large language model (LLM) to find up to $n_s$ alternatives to the user prompt. 
This search early stops once an alternative prompt \textcolor{red}{$e_{p}^{ss}$}'s magnitude falls below the $\lambda$ threshold. Visually in \cref{fig:method} (c), this is equivalent to reaching the target privacy contour at points \texttt{B'}, \texttt{C'}, or \texttt{D'}.
Notably, although \texttt{B} and \texttt{B'} lie on the same privacy contour, the utility of \texttt{B'} is often higher than that of \texttt{B}.
Implementationally, if all $n_s$ alternatives result in magnitudes above $\lambda$, the one with the lowest magnitude will be chosen.

\textbf{Analysis: Enhancing Utility with No Privacy Cost.}
This semantic search differs from soft prompt engineering in that the optimization happens in the language space instead of the embedding space.
The baseline strategy seeks to boost utility by minimizing the number of prompt optimization steps, which diminishes the extent of prompt modifications, significantly undermining privacy as ostensibly secure prompts may inadvertently produce memorized outputs.
Conversely, our approach employs a semantic prompt search to inherently preserve utility, leveraging the capabilities of Large Language Models (LLM) to navigate the linguistic space for suitable alternatives. 
By pinpointing a substitute prompt in the language space, we keep the semantic similarity and obviate the need to trade privacy for enhanced utility.

\textbf{Analysis: Enhancing Privacy with Minimum Utility Cost}. 
This is also guaranteed by design.
As discussed in \cref{sec:np}, the baseline requires significant utility sacrifices to gain extra privacy.
In contrast, the utility impact associated with our semantic prompt search is limited to the semantic discrepancy between the original user prompt and the LLM-derived alternative, which is much smaller.
Furthermore, when the search yields an alternative prompt devoid of memorization risks, it effectively represents a significant boost in privacy for a minimal cost of utility.
Geometrically as in \cref{fig:method} (c), this means shifting the prediction from \texttt{O} to \texttt{D'} with minimum utility cost.

\begin{figure}[tb]
  \centering
  \includegraphics[height=4.cm]{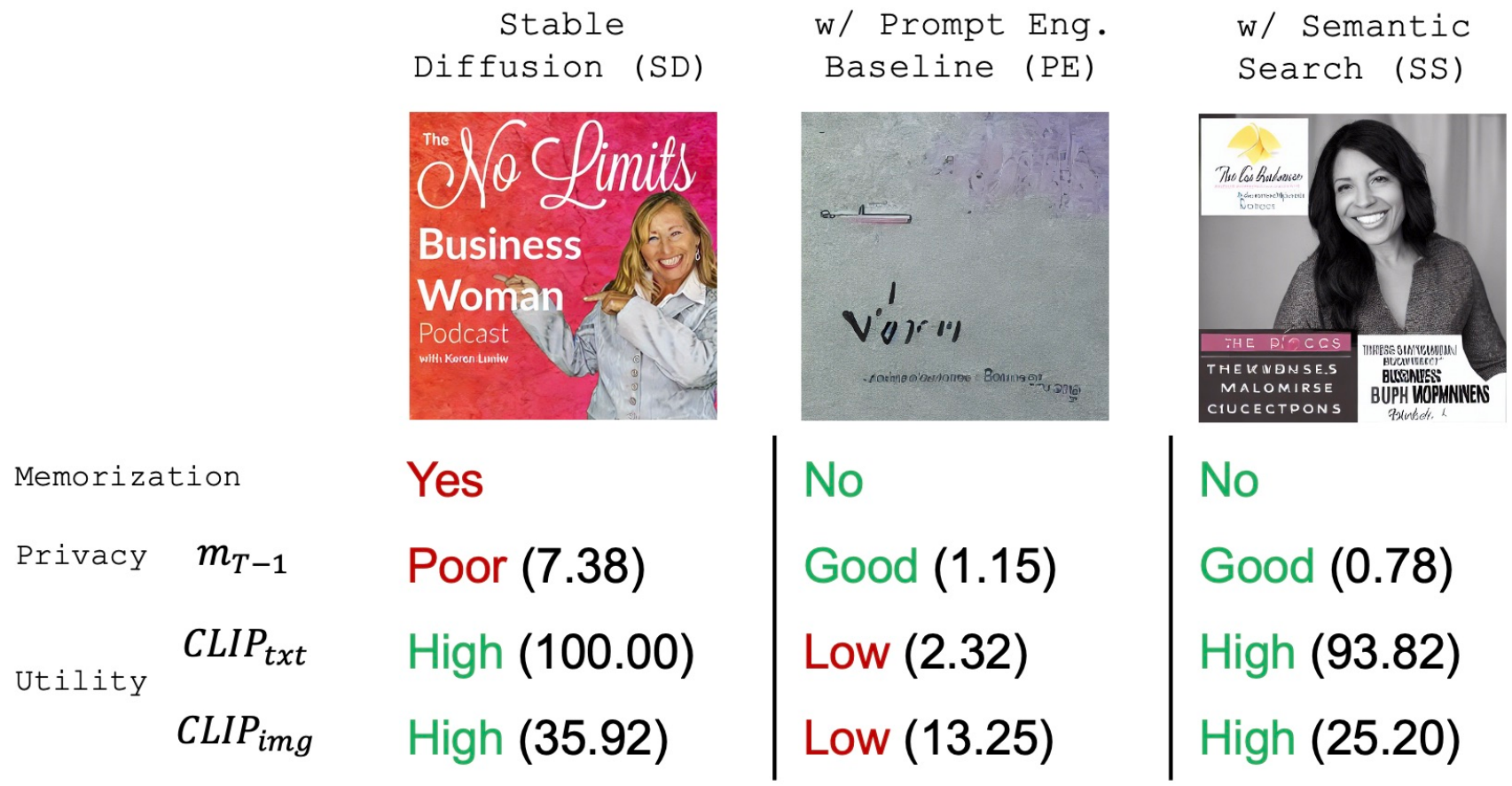}
  \caption{Both the baseline approach and our method effectively inhibit Stable Diffusion from producing memorized images by minimizing the magnitude in the initial inference step for prompt ``The No Limits Business Woman Podcast". Nonetheless, our \textit{semantic prompt search (SS)} strategy, which results in ``The Empowered Business Woman’s Podcast", retains considerably more utility, as demonstrated by the substantially higher CLIP similarities between the original user text prompt and both our (1) modified prompt, and (2) ultimately generated image. Qualitative results further corroborate this finding. $m_{T-1}$ is the first-step magnitude, $CLIP_{txt}$ and $CLIP_{img}$ refer to the original prompt's CLIP similarity with the modified prompt and the generated image.
  }
  \label{fig:analysis_ss}
\vspace{-0.6cm}
\end{figure}

\cref{fig:analysis_ss} demonstrate SS's superiority over the baseline using a widely-investigated memorized prompt: ``The No Limits Business Woman Podcast". 
While the baseline’s prompt engineering (PE) strategy successfully reduces the magnitude to 1.15 and prevents memorization, it results in a significant semantic deviation from the original prompt, as indicated by low CLIP similarity scores for both the engineered text prompt and the generated image. In contrast, our SS strategy not only reduces the magnitude further to 0.78, effectively preventing memorization, but also significantly improves text alignment. By navigating the linguistic space, our method finds ``The Empowered Business Woman's Podcast" as a privacy-preserving yet semantically aligned alternative.

\subsection{Synergy Effect of the Two Strategies}
\label{sec:synergy}
\vspace{-0.cm}

\textbf{Analysis: Tackling the Limitation of Semantic Search (SS)}.
In an extreme scenario, SS might encounter a limitation where all $n_s$ searched alternatives result in magnitudes exceeding the set threshold, failing to meet the required privacy standards. This situation can be geometrically represented by a movement from point \texttt{O} to \texttt{B'} in \cref{fig:method} (c). However, it's important to note that even in such cases, the transition to a lower magnitude contour, where \texttt{B'} is located, still offers a certain degree of privacy enhancement without additional cost. Further advancements in privacy (\eg, moving from \texttt{B'} to \texttt{D'}) can be achieved through our synergistic application of the PR strategy, which incurs a smaller utility cost compared to the baseline method.
The right side of \cref{fig:analysis_pr_synergy} (in Supplementary Material) showcases the synergistic benefit of utilizing the PR strategy to compensate for the SS strategy's limitations. 
In this example, although SS may not secure an alternative prompt with a sufficiently reduced magnitude, lowering it from the original 7.48 to 6.02 has proved to be adequate for PR to leverage the original prompt as the new anchor point for steering the generation away from memorized content effectively.

\textbf{Analysis: Addressing the Privacy-Utility Dilemma}.
For the baseline strategy \cite{wen_2024_iclr}, the dilemma in balancing the privacy-utility trade-off through adjusting the $\lambda$  hyperparameter manifests in two ways: opting for a higher $\lambda$ to bolster utility inadvertently raises the False Negative (FN) rate, leading to potentially unsafe prompts being misclassified as safe and thus compromising privacy; conversely, selecting a lower $\lambda$ to enhance privacy increases the False Positive (FP) rate, resulting in unnecessary modifications to already safe prompts, thereby reducing utility.
\textit{Prompt re-anchoring} is tailored to mitigate the privacy repercussions of False Negatives (FN) encountered when users aim to augment utility with a higher $\lambda$. It achieves this by actively directing generation away from memorized content, as elucidated in \cref{sec:np} and depicted in \cref{fig:method} as a divergence from \texttt{O}. This approach thus optimizes the privacy-utility trade-off under such conditions.
\textit{Semantic prompt search}, on the other hand, aims to alleviate the utility decline tied to False Positives (FP) when enhancing privacy with a lower $\lambda$. It conducts a linguistically sensitive search that avoids semantic degradation, consequently refining the privacy-utility balance in these scenarios.
Employing both strategies synergistically ensures an enhanced privacy-utility trade-off, regardless of whether the user's priority lies with maximizing privacy or utility. 

\section{Experiments}
\subsection{Evaluation settings} 
\label{sec:settings}
Although our method is designed to enhance the approach in~\cite{wen_2024_iclr}, it can be easily adapted to any recent advancement of the detection strategy by simply replacing $m_{T-1}$ in \cref{eq:gpt} with alternative detection signals, making PRSS highly practical and adaptable for future improvements. For example, a contemporary work~\cite{chen_be} introduced the masked magnitude $m'_{t}$ (see \cref{eq:masked_magnitude}), achieving new state-of-the-art detection performance. 
The mitigation strategy proposed by~\cite{chen_be} also uses prompt engineering based on the upgraded detection signal $m'_{t}$, offering two key advantages: (1) $m'_{t}$ provides a more accurate detection trigger for initiating the mitigation strategy, (2) it serves as a more effective loss function than unmasked magnitude, yielding better gradients for optimizing prompts. These improvements make it the current state-of-the-art in mitigation.
PRSS can seamlessly integrate this enhanced detection signal by replacing $m_{T-1}$ with $m'_{T-1}$ in \cref{eq:gpt}, producing our best results to date.
For a fair comparison, we re-implement the mitigation strategy from~\cite{chen_be} and evaluate its performance alongside PRSS. This comparison not only highlights PRSS's capability to surpass the most recent state-of-the-art but also serves as an ablation study, assessing PRSS’s impact since both methods use the same detection strategy to trigger mitigation.
Additionally, we produce a version of PRSS that uses $m_t$ as the detection signal, aligning with~\cite{wen_2024_iclr}, to further demonstrate PRSS’s effectiveness in comparison to~\cite{wen_2024_iclr}.

We follow the baselines~\cite{wen_2024_iclr, chen_be} by experimenting on Stable Diffusion v1-4, using its corresponding prompt dataset from~\cite{Webster2023}, which contains 500 prompts, over 300 of which are classified as memorized and organized by different memorization types. To provide a comprehensive view, we adopt the settings from~\cite{chen_be} and report results for both global and local memorization prompts separately.
We use the default hyperparameters from~\cite{wen_2024_iclr} to re-implement the detection and prompt engineering mitigation strategies from~\cite{wen_2024_iclr} and~\cite{chen_be}.
For semantic prompt search with LLMs, we use the GPT-4 model via the OpenAI API, instructing it to generate up to $n_s=25$ semantically similar prompts for each user prompt. The associated inference time and cost are minimal, with each alternative prompt generated in approximately 0.9 seconds, costing around \$0.02.

Following standard practice in memorization literature, we use the CLIP score to measure the alignment between the generated output and the input user prompt, serving as our utility metric. For privacy metrics, which quantify the degree of memorization, most studies rely on SSCD-based scores (see \cref{eq:sscd}), reporting metrics such as the average score, the 95th percentile score, and the percentage of memorized generations, with a threshold of 0.5 commonly indicating memorization.
Additionally, \cite{chen_be} proposes a localized similarity metric (LS, see \cref{eq:ls}) that measures similarity between the generated image and corresponding training image only within the localized memorization region. 
We present comparative results using the widely adopted SSCD metric in \cref{fig:main} and \cref{fig:ablation}, while other metrics, including LS, are detailed in the Supplementary Material for a more comprehensive evaluation.

\subsection{Comparison with Baselines}
\begin{figure}[tb]
  \centering
  \includegraphics[width=1.0\linewidth]{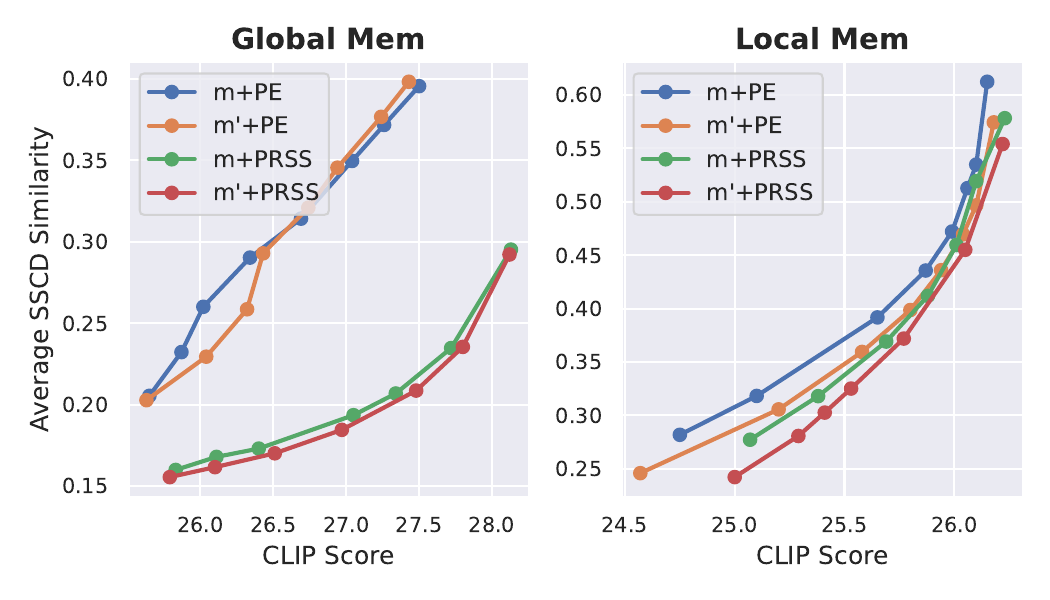}
  \caption{Comparison with baselines. Across different detection signals ($m$ and $m'$), PRSS consistently achieves superior privacy-utility trade-offs. The improvement is particularly pronounced in global memorization compared to local memorization.}
  \label{fig:main}
\vspace{-0.5cm}
\end{figure}

As shown in \cref{fig:main}, PRSS consistently outperforms prompt engineering (PE) in achieving better privacy-utility trade-offs, regardless of the detection signal used (whether $m$ or $m'$). This is visually evident as the red and green lines (PRSS) lie further in the bottom-right direction compared to the yellow and blue lines (PE) for both global and local memorization scenarios.
Notably, PRSS demonstrates astonishingly large improvements in global memorization (from the blue line to the green line and from the yellow line to the red line), while \cite{chen_be} achieves only marginal gains over \cite{wen_2024_iclr} in this case (from the blue line to the yellow line and from the green line to the red line).
For local memorization, although the margin of improvement is smaller, PRSS still achieves a significant improvement over PE. 
Specifically, even using the inferior detection signal $m$, PRSS (green line) achieves comparable performance with the method proposed by \cite{chen_be} (yellow line), which leverages the superior signal $m'$.
Equipped with $m'$, PRSS can achieve the state-of-the-art results (red line). This highlights PRSS’s robust capability to leverage advancements in detection signals.

\subsection{Ablation Studies}
\begin{figure}[tb]
  \centering
  \includegraphics[width=1.0\linewidth]{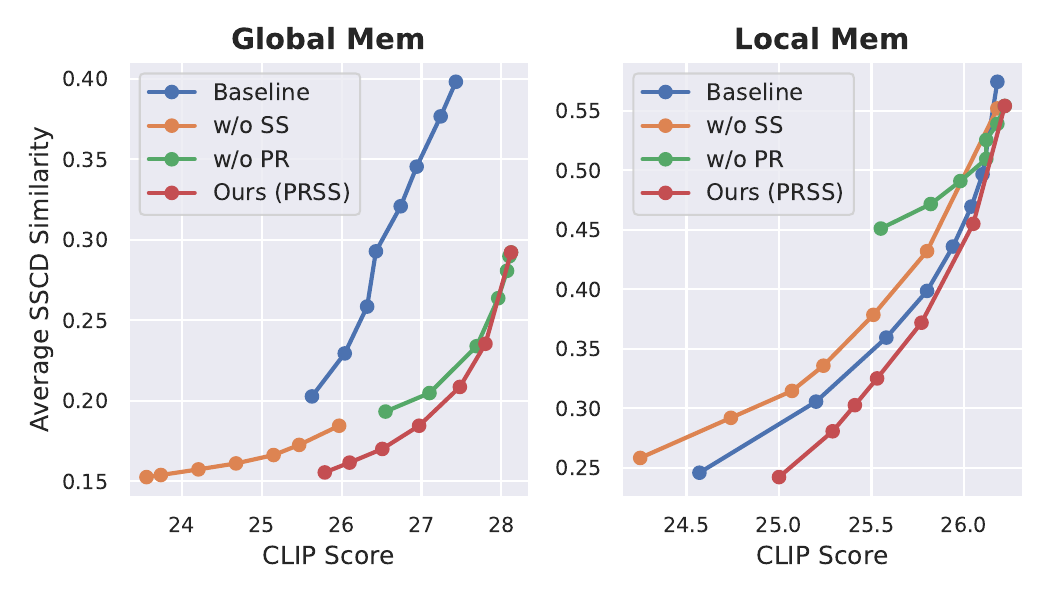}
  \vspace{-0.8cm}
  \caption{Ablation studies of PRSS components. Using PR alone results in low utility, while SS alone cannot ensure privacy. Combining PR and SS achieves a synergistic effect, delivering the best privacy-utility trade-off.}
  \label{fig:ablation}
\vspace{-0.5cm}
\end{figure}

\cref{fig:ablation} highlights the necessity of both components in PRSS, as evidenced by the red line lying further in the bottom-right direction compared to the yellow and green lines for both global and local memorization scenarios.
The synergistic effect of combining PR and SS as discussed in \cref{sec:synergy}, is particularly evident in the local memorization scenario, which aligns with \cite{chen_be}'s observation that local memorization poses greater challenges for existing mitigation strategies. Using PR alone effectively enhances privacy but results in suboptimal utility, while SS alone improves utility but achieves limited privacy protection. Together, they leverage each other's strengths, achieving the best privacy-utility trade-off.
For global memorization, either PR or SS alone significantly outperforms the baseline. However, their combination further enhances the trade-off, underscoring the robustness of PRSS in tackling both global and local memorization challenges.

\vspace{-0.3cm}
\section{Related Work}
\vspace{-0.2cm}
\label{sec:related}

\textbf{Memorization in Generative Models} 
The phenomenon of generative models memorizing training data has been extensively studied across domains, including language models~\cite{12, 44, 45, 51, carlini2022quantifying, related_general}, image generation~\cite{74, 3, 36, related_general_2, related_general_3, related_general_5}, and video generation~\cite{framebyframe, mem_other_modality_video}, due to the significant privacy and legal risks it poses~\cite{privacy_concerns_1, privacy_concerns_2, privacy_concerns_3, lawsuit}.
Specifically, in diffusion models, \cite{somepalli_2023_cvpr, carlini_2023_usenix, Webster2023} first validated the memorization issue on various models including DDPMs \cite{ddpm, iddpm_2021_icml} and Stable Diffusion \cite{sd_2022_cvpr} on datasets include Oxford Flowers \cite{oxfordflowers}, CelebA \cite{celeba}, CIFAR-10 \cite{cifar10} and LAION \cite{LAION5B}. 
Further analysis by \cite{somepalli_2023_neurips} revealed that Stable Diffusion’s text-conditioning exacerbates overfitting, making its memorization of LAION particularly problematic.  
To better understand memorization, \cite{Webster2023} categorized the phenomenon as Matching Verbatim, Retrieval Verbatim, and Template Verbatim based on the cause, distinguishing between one-to-one and many-to-many mappings in text-image pairs, while \cite{chen_be} classified memorization into global and local cases, demonstrating the practical significance of this distinction.  
In this work, we focus on Stable Diffusion due to its widespread use, open-source nature, and real-world legal challenges~\cite{lawsuit}. We adopt the framework of \cite{chen_be} to analyze both global and local memorization for a more nuanced evaluation of our proposed strategies.

\textbf{Detection Strategies for Diffusion Memorization.}
Early detection methods required generating large image sets to estimate density~\cite{carlini_2023_usenix} or comparing generations with the entire training dataset~\cite{somepalli_2023_cvpr}, which were computationally intensive. Recent approaches~\cite{ren2024unveiling, wen_2024_iclr} introduced more efficient one-step detection strategies.  
\cite{ren2024unveiling} detected memorization by observing less concentrated cross-attention scores for memorized generations than non-memorized ones. \cite{wen_2024_iclr} proposed using the abnormally high magnitude of text-conditional noise predictions as a detection signal. Building on this, \cite{chen_be} incorporated localized memorization masks into the magnitude computation, offering a more accurate detection signal and achieving state-of-the-art performance.  
Our proposed PRSS approach leverages these detection mechanisms to trigger mitigation strategies. Notably, PRSS can adapt to any detection method by substituting the indicator variable condition in its framework. We employ the state-of-the-art method from \cite{chen_be} for optimal results.

\textbf{Mitigation Strategies for Diffusion Memorization.}  
Mitigation strategies have evolved rapidly. Early methods, such as retraining models on de-duplicated data~\cite{carlini_2023_usenix}, were computationally expensive and showed limited effectiveness. ~\cite{somepalli_2023_neurips} introduced randomness into text prompts or embeddings but indiscriminately affected non-memorized prompts, reducing overall utility.  
To improve on it, \cite{chen-amg} proposed to compute the generation's similarity with the closest training image to engineer the image latent using the classifier guidance framework. However, this relies on computationally expensive nearest-neighbor searches across the training dataset. 
\cite{wen_2024_iclr} proposed to conduct prompt engineering using the magnitude of text-conditional prediction as the loss, and \cite{ren2024unveiling} proposed to re-weight cross-attention scores for all tokens in the prompt. They offer improved utility-privacy trade-offs and achieve comparable performance.
Most recently, \cite{chen_be} extended these efforts by leveraging localized magnitudes as loss functions for prompt engineering, achieving the best results to date.  
Our investigation reveals the limitations in the prevailing method's privacy-utility balance. In response, we introduce new mitigation strategies that achieve more favorable trade-offs.

\vspace{-0.2cm}
\section{Conclusion and Future Work}
\vspace{-0.2cm}

This paper introduces a novel \textit{PRSS} method that consists of \textit{prompt re-anchoring} and \textit{semantic prompt search} as deliberately designed strategies to mitigate memorization in diffusion models. 
Empirical results show that PRSS demonstrates superior privacy-utility trade-offs over existing methods.
Beyond its effectiveness, PRSS is simple and efficient to implement. It requires only minor modifications to the CFG equation to update noise predictions during inference, seamlessly integrating into the inference pipeline without additional training, fine-tuning, or reliance on training data. The efficacy of each component is validated through comprehensive analysis and ablation studies.
A current limitation is that the performance of PRSS, as well as baseline mitigation strategies, relies heavily on the accuracy of the detection mechanism. When it fails to trigger accurately, they all revert to the standard Stable Diffusion process, compromising privacy preservation. Thus, improving the detection mechanism's accuracy in future works would orthogonally bolster the effectiveness of PRSS.

\section{Acknowledgement}
This work was supported in part by the Australian Research Council under Projects DP210101859 and FT230100549. 
{
    \small
    \bibliographystyle{ieeenat_fullname}
    \bibliography{main}
}
\clearpage
\setcounter{page}{1}
\maketitlesupplementary
\paragraph{Outline.}
In \cref{sec:1}, we provide additional analysis explaining why the \textit{prompt re-anchoring (PR)} strategy effectively enhances privacy and how synergistic benefits emerge when combined with the \textit{semantic search (SS)} strategy.
In \cref{sec:2}, we present additional quantitative results, showing that the proposed PRSS strategy consistently improves the privacy-utility trade-offs compared to baseline methods. These improvements are demonstrated under different detection signals for both global and local memorization scenarios and evaluated using various memorization metrics.
In \cref{sec:3}, we include additional qualitative results highlighting PRSS's superior performance over baseline approaches in improving privacy while maintaining high utility.
In \cref{sec:4}, we delve into further implementation details to enhance the reproducibility of our work.
\section{Additional Analysis}
\label{sec:1}
\begin{figure*}[tb]
  \centering
  \includegraphics[width=\textwidth]{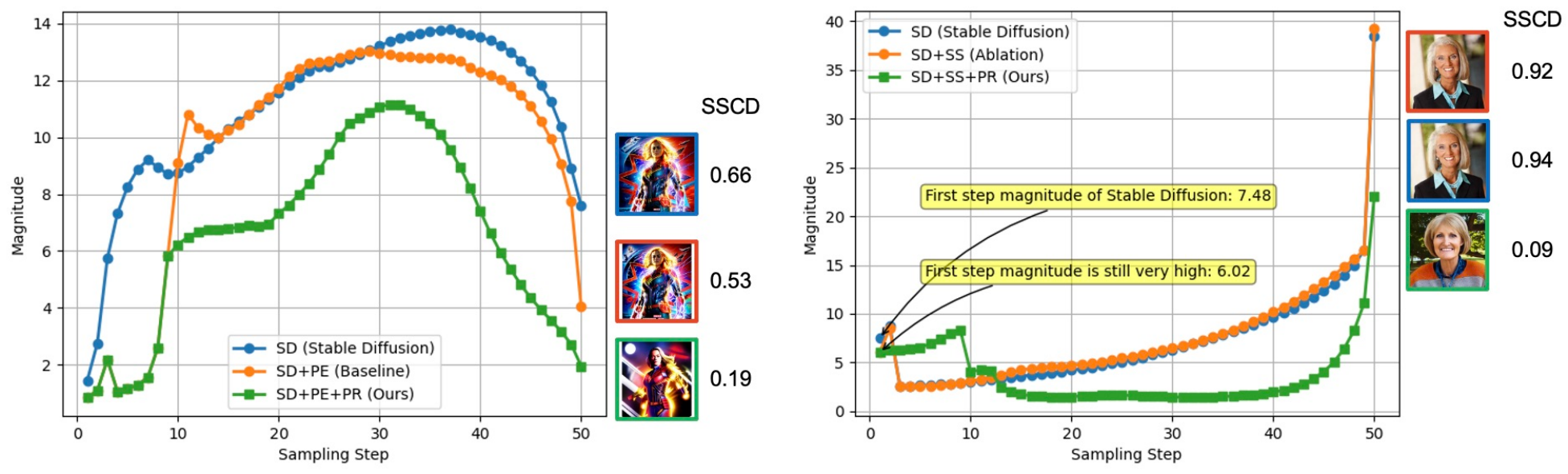}
  \caption{\textbf{Left and Right}: Our \textit{prompt re-anchoring (PR)} strategy allows continuous memorization diversion throughout the entire inference process, effectively reducing magnitude and enhancing privacy. \textbf{Right}: Should the alternative prompt identified by the \textit{semantic prompt search (SS)} strategy not achieve a low magnitude, employing \textit{PR} concurrently can still prevent memorized image generation, provided the magnitude of the alternative prompt is lower than that of the original.}
  \label{fig:analysis_pr_synergy}
\vspace{-0.6cm}
\end{figure*}

\cref{fig:analysis_pr_synergy} demonstrates the shortcomings of the baseline's prompt engineering strategy in preserving privacy and the advantages of our proposed \textit{prompt re-anchoring (PR)} and \textit{semantic search (SS)} strategies. 
Please check out the main paper for a detailed analysis, the key takeaways are summarized as follows.

\paragraph{Enhancing Privacy Through PR.}
As shown in both line charts in \cref{fig:analysis_pr_synergy}, while the engineered prompt achieves a low magnitude initially, it fails to ensure that this low level is maintained throughout subsequent inference steps, potentially leading back to high magnitudes indicative of Stable Diffusion's memorized outputs.
In response, our \textit{PR} strategy introduces a mechanism for consistent diversion from memorization across the entire inference process. This is demonstrated by consistently lower magnitudes and corroborated by both the qualitative appearance of the final generated image and its quantitative similarity score, showcasing effective privacy preservation.

\paragraph{Synergy Effect of the Two Strategies.}
The right side of \cref{fig:analysis_pr_synergy} showcases the synergistic benefit of utilizing the \textit{prompt re-anchoring} strategy to compensate for the \textit{SS} strategy's limitations. 
Although the semantic search may not secure an alternative prompt with a sufficiently reduced magnitude, lowering it from the original 7.48 to 6.02 is proved to be adequate for \textit{prompt re-anchoring} to leverage the original prompt as the new anchor point for steering the generation away from memorized content effectively.

\section{Additional Quantitative Results}
\label{sec:2}
\begin{figure}[tb]
  \centering
  \includegraphics[width=1.0\linewidth]{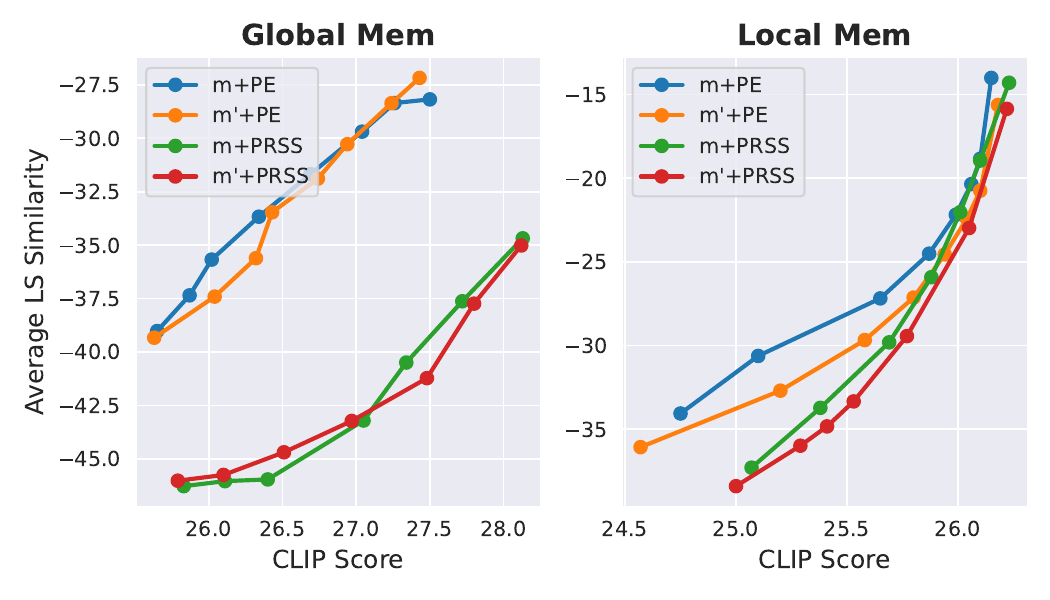}
  \caption{Comparison with baselines using localized memorization metric LS. Across different detection signals ($m$ and $m'$), PRSS consistently achieves superior privacy-utility trade-offs.}
  \label{fig:main_ls}
\end{figure}

\begin{figure}[tb]
  \centering
  \includegraphics[width=1.0\linewidth]{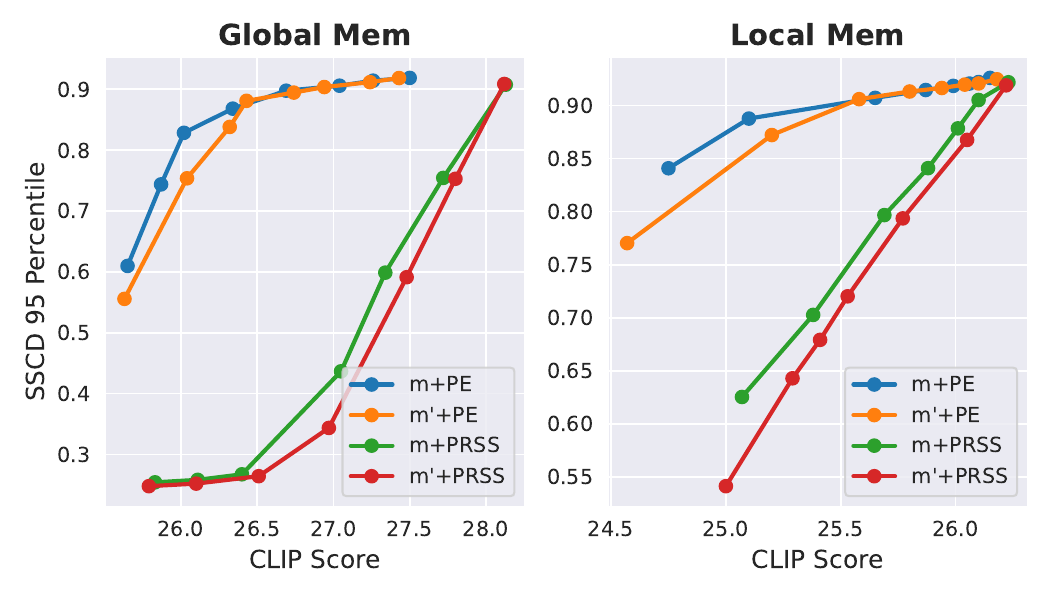}
  \caption{Comparison with baselines using the 95 percentile of SSCD similarity scores as the memorization metric. Across different detection signals ($m$ and $m'$), PRSS consistently achieves superior privacy-utility trade-offs.}
  \label{fig:main_95}
\end{figure}

\begin{figure}[tb]
  \centering
  \includegraphics[width=1.0\linewidth]{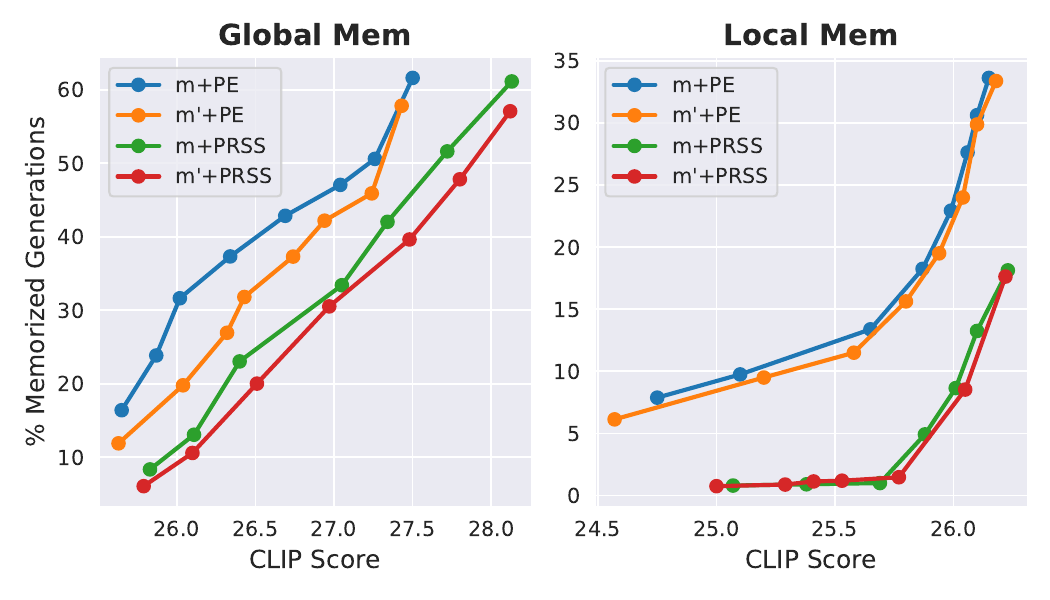}
  \caption{Comparison with baselines using the percentage of memorized generations (quantified as SSCD similarity $> 0.50$) as the memorization metric. Across different detection signals ($m$ and $m'$), PRSS consistently achieves superior privacy-utility trade-offs.}
  \label{fig:main_perc}
\end{figure}

As discussed in \cref{sec:settings}, we complement the standard average SSCD similarity score that is reported in \cref{fig:main} with additional metrics for a comprehensive evaluation. 
These include the localized similarity metric LS (\cref{eq:ls}), designed specifically for assessing local memorization~\cite{chen_be}, shown in \cref{fig:main_ls}.
We also report other SSCD-based statistics, such as the 95th percentile similarity score used in~\cite{somepalli_2023_neurips} (\cref{fig:main_95}) and the percentage of memorized generations (SSCD similarity $> 0.50$) employed in~\cite{carlini_2023_usenix}, as depicted in \cref{fig:main_perc}.

Same as \cref{fig:main}, these additional results are presented separately for global and local memorization scenarios, comparing PRSS against two baselines (\cite{wen_2024_iclr} and \cite{chen_be}) using their respective detection signals to ensure fair comparisons. 

Across all metrics, the results confirm that PRSS consistently improves the privacy-utility trade-offs of both baselines, as evidenced by the red line consistently lying in the bottom-right direction of the yellow line, and the green line consistently lying in the bottom-right direction of the blue line for both global and local memorization scenarios.

\section{Additional Qualitative Results}
\label{sec:3}
\begin{figure*}[h]
  \centering
  \includegraphics[width=\textwidth]{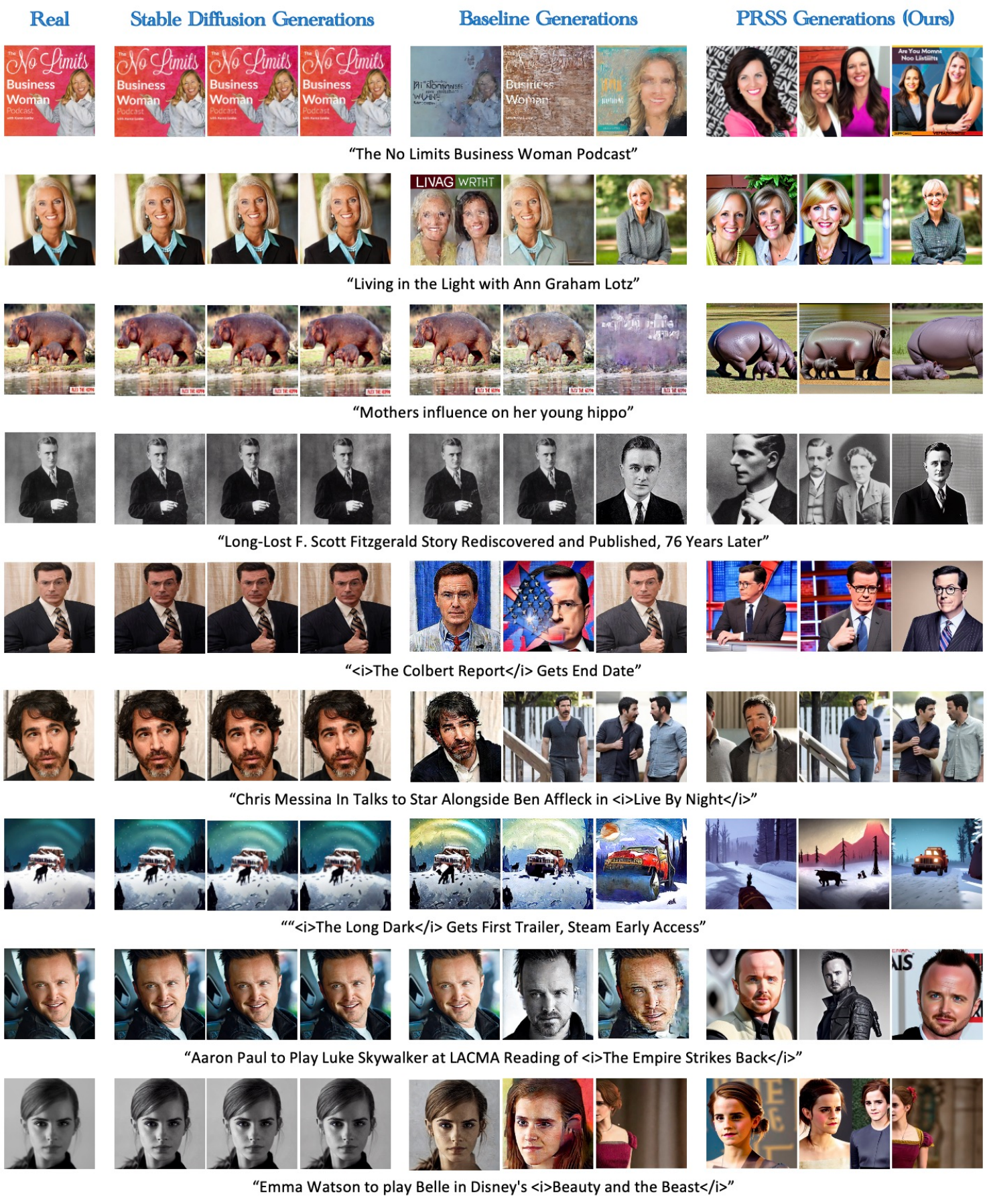}
  \vspace{-0.7cm}
  \caption{Qualitative comparisons for global memorization using the same random seeds. The baseline's prompt engineering approach often incurs a significant loss of output utility or fails to adequately mitigate memorization. PRSS effectively reduces memorization in Stable Diffusion while maintaining semantic alignment between the generated images and the user's prompt.}
  \label{fig:aditional1}
\end{figure*}

\begin{figure*}[h]
  \centering
  \includegraphics[width=\textwidth]{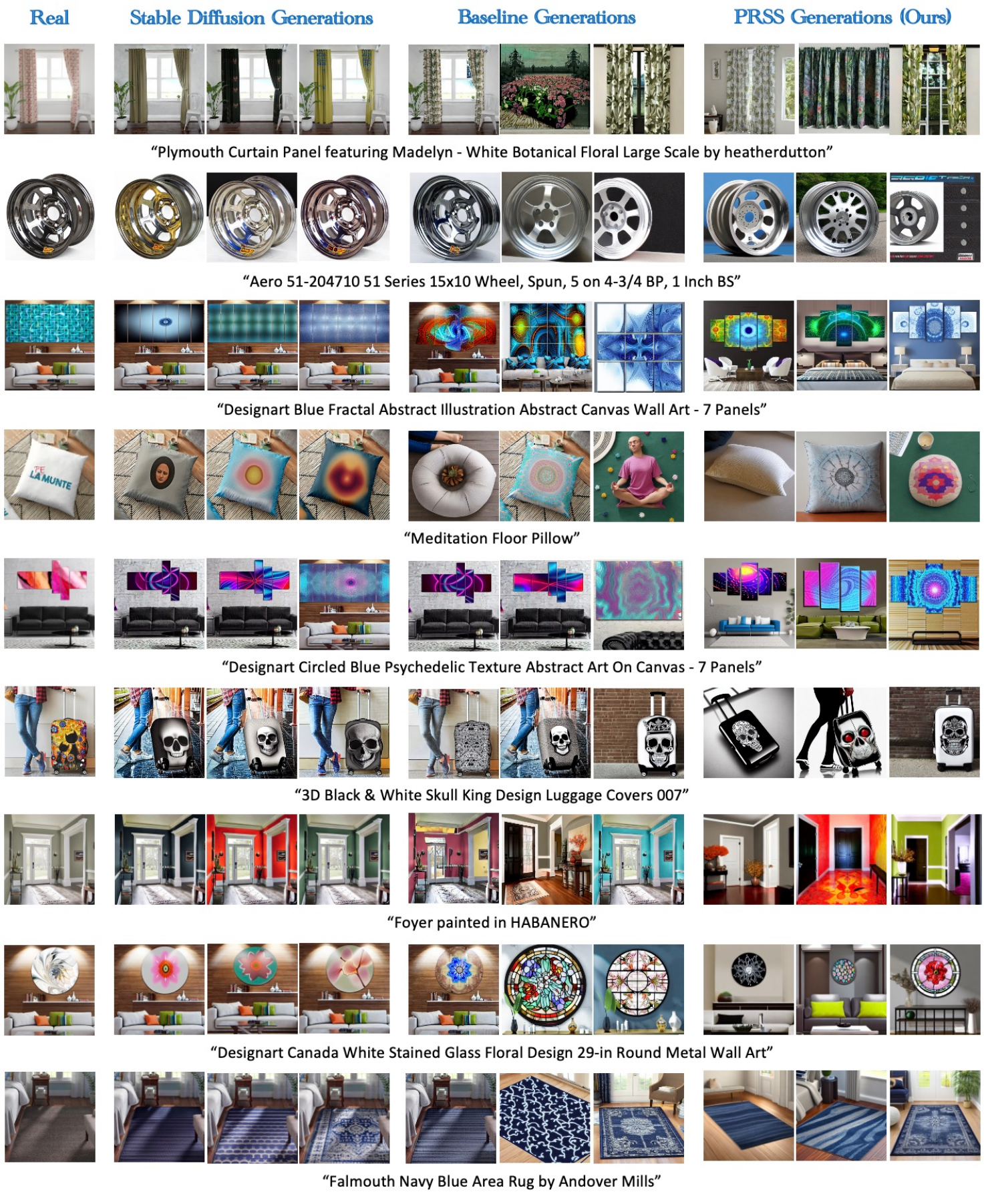}
  \vspace{-0.8cm}
  \caption{Qualitative comparisons for local memorization using the same random seeds. The baseline's prompt engineering approach often incurs a significant loss of output utility or fails to adequately mitigate memorization. PRSS effectively reduces memorization in Stable Diffusion while maintaining semantic alignment between the generated images and the user's prompt.}
  \label{fig:aditional2}
\end{figure*}

We present additional qualitative results in \cref{fig:aditional1,fig:aditional2} for comparative analysis in both local and global memorization scenarios using the same random seeds. These results demonstrate the superior performance of our method in enhancing privacy while preserving utility, effectively reducing memorization within Stable Diffusion generations.

Specifically, generations produced by baseline strategies that leverage prompt engineering often exhibit poor alignment with the intended meaning of the user prompt or continue to replicate training images similar to those generated by the original Stable Diffusion model. This highlights their limitations in maintaining both utility and privacy. In contrast, our approach produces images that are significantly less prone to memorization while remaining semantically aligned with the user's prompt.

\section{Additional Implementational Details}
\label{sec:4}

\subsection{Controlling the Strength of PR}
We achieve finer-grained control of the privacy-utility trade-off by adjusting the strength of PR guidance. Specifically, we balance the guidance scale $s$ between fully using the re-anchored guidance method (\cref{eq:prss_all_pr}) and fully using the original guidance method (\cref{eq:prss_all_null}). 
\begin{equation}
\begin{split}
\hat{\epsilon} \leftarrow [{\epsilon}_{\theta}(x_t, e_{\phi}) + s ({\epsilon}_{\theta}(x_t, \textcolor{brown}{e_p}) - {\epsilon}_{\theta}(x_t, e_{\phi}))] \mathbbm{1}_{\{m_{T-1} < \lambda\}} \\
+ [{\epsilon}_{\theta}(x_t, \textcolor{brown}{e_p}) + s ({\epsilon}_{\theta}(x_t, \textcolor{red}{e_{p}^{ss}}) - {\epsilon}_{\theta}(x_t, \textcolor{brown}{e_p}))] \mathbbm{1}_{\{m_{T-1} > \lambda\}}.
  \label{eq:prss_all_pr}
\end{split}
\end{equation}
\begin{equation}
\begin{split}
\hat{\epsilon} \leftarrow [{\epsilon}_{\theta}(x_t, e_{\phi}) + s ({\epsilon}_{\theta}(x_t, \textcolor{brown}{e_p}) - {\epsilon}_{\theta}(x_t, e_{\phi}))] \mathbbm{1}_{\{m_{T-1} < \lambda\}} \\
+ [{\epsilon}_{\theta}(x_t, e_{\phi}) + s ({\epsilon}_{\theta}(x_t, \textcolor{red}{e_{p}^{ss}}) - {\epsilon}_{\theta}(x_t, e_{\phi}))] \mathbbm{1}_{\{m_{T-1} > \lambda\}}.
  \label{eq:prss_all_null}
\end{split}
\end{equation}
The balanced approach, shown in \cref{eq:prss_balanced}, splits the guidance scale $s$ between these two options, allowing for a more flexible trade-off.
\begin{equation}
\begin{split}
\hat{\epsilon} \leftarrow [{\epsilon}_{\theta}(x_t, e_{\phi}) + s ({\epsilon}_{\theta}(x_t, \textcolor{brown}{e_p}) - {\epsilon}_{\theta}(x_t, e_{\phi}))] \mathbbm{1}_{\{m_{T-1} < \lambda\}} \\
+ [{\epsilon}_{\theta}(x_t, \textcolor{brown}{e_p}) + s' ({\epsilon}_{\theta}(x_t, \textcolor{red}{e_{p}^{ss}}) - {\epsilon}_{\theta}(x_t, \textcolor{brown}{e_p})) + 
\\
(s-s') ({\epsilon}_{\theta}(x_t, \textcolor{red}{e_{p}^{ss}}) - {\epsilon}_{\theta}(x_t, e_{\phi}))] \mathbbm{1}_{\{m_{T-1} > \lambda\}}.
  \label{eq:prss_balanced}
\end{split}
\end{equation}
where $s' \in [1, s]$: when $s' = 1$, the method defaults to the original guidance using $e_{\phi}$ as the anchor; when $s' = s$, the guidance term is fully re-anchored to \textcolor{brown}{$e_p$}. These represent the two extreme cases.
Specifically, since the magnitude quantifies the level of memorization risk, we design $s'$ to be proportional to the real-time magnitude at timestep $t$ throughout the entire inference process. It is capped at $s$ for excessively large magnitudes that are greater than $\lambda_{\text{max}}$ and floored at $1$ when the magnitude is below the detection threshold $\lambda$, ensuring no PR is triggered during non-memorized cases to preserve output utility maximally. This behavior is governed by the following specifications:
\begin{equation}
s' = 1 + (s - 1) \cdot \left( \frac{\max(\min(m_{t}, \lambda_{\text{max}}) - \lambda, 0)}{\lambda_{\text{max}} - \lambda} \right)
\label{eq:scale}
\end{equation}

\subsection{Generating Alternative Prompts}
As discussed in \cref{sec:settings}, we leverage GPT-4 via the OpenAI API for searching semantically similar alternative text prompts. 
To ensure reproducibility, we describe the process we used for such a task.
After taking a user-provided text prompt, the process begins by setting up a system instruction that explicitly directs GPT-4 to act as a skilled prompt engineer. The instruction focuses on paraphrasing the input prompt into semantically similar, natural-sounding alternatives. The exact instruction provided to GPT-4 is as follows:
\textit{``You are a prompt engineer skilled in paraphrasing user prompts into semantically similar, natural-sounding prompts. I will provide a text prompt that will be fed into Stable Diffusion. Please generate 1 semantically similar prompt as a plain sentence, without using any list numbering (such as 1., 2., etc.) or quotation marks. Each prompt should be a standalone sentence in plain language."}

The OpenAI API is configured to query the GPT-4 model with these messages. Parameters such as \texttt{max\_tokens}, \texttt{temperature}, and the number of alternatives (\texttt{n}) are adjusted to optimize the generation process. For instance, \texttt{max\_tokens=750} ensures sufficient response length, \texttt{temperature=0.8} introduces slight randomness for diversity, and \texttt{n=1} generates a single prompt per call for focused evaluation.

Below, we list some examples of generated alternative prompts (in bullet points) for a given user text prompts (in bold):

\vspace{0.3cm}
\noindent \textbf{\texttt{The No Limits Business Woman Podcast}}
\begin{itemize}
    \item The Businesswoman Without Boundaries Show
    \item The Entrepreneurial Woman's Unrestricted Podcast
    \item Podcast for the Unstoppable Business Lady
    \item The Infinite Potential Business Woman Series
    \item The Female Mogul's Limitless Journey Podcast
    \item Broadcast for the Fearless Businesswoman
    \item The Ceaseless Enterprise Woman Podcast
    \item Unbounded Success: A Woman's Business Podcast
    \item The Entrepreneurial Woman's No-Barriers Broadcast
    \item The Female Executive's Infinite Podcast
    \item The Business Femme's Unchained Podcast Series
    \item Undeterred: The Business Woman's Podcast
    \item The Unrestricted Businesswoman's Dialogue
    \item The Empowered Business Woman's Podcast
    \item Ventures Without Limits: A Female Entrepreneur's Podcast
    \item The Business Woman's Sky's-the-Limit Show
    \item Limitless Horizons: The Woman Entrepreneur Podcast
    \item The Trailblazing Businesswoman's Podcast
    \item Podcasting the Unstoppable Business Woman
    \item Frontiers of Business: The Woman's Edition
    \item The Entrepreneurial Spirit: Women Without Limits
    \item The Podcast for Limit-Defying Businesswomen
    \item Beyond Barriers: The Female Business Leader Podcast
    \item The Businesswoman's Podcast: No Boundaries
    \item Empire Building: The Woman's Podcast without Limit
\end{itemize}
\vspace{0.3cm}

\noindent \textbf{\texttt{Mothers influence on her young hippo}}
\begin{itemize}
    \item Mother hippo's guidance of her offspring
    \item The impact of a mother on her baby hippo
    \item Maternal effect on a juvenile hippo
    \item The role of a mother in a young hippo's life
    \item Motherly impact on a young hippopotamus
    \item The nurturing of a baby hippo by its mother
    \item Influence of a mother hippo on her calf
    \item How a mother shapes her hippo youngster
    \item Maternal shaping of a young hippo's development
    \item The nurturing bond between mother and baby hippo
    \item Maternal teachings to a young hippopotamus
    \item The guiding presence of a mother in her hippo's upbringing
    \item A mother hippo's role in her offspring's growth
    \item The maternal touch in raising a young hippo
    \item Motherly wisdom bestowed upon a juvenile hippo
    \item A mother's guiding influence on her young hippopotamus
    \item Life lessons from a mother to her hippo calf
    \item The formative influence of a mother on her baby hippo
    \item A hippo mother's impact on her offspring's early life
    \item Maternal instincts shaping a young hippo's journey
    \item The protective guidance of a mother hippo to her young
    \item Young hippo's upbringing under its mother's care
    \item The maternal effect on a hippo calf's behavior
    \item Influence of motherhood on a young hippo's path
    \item Mother and young hippo: A nurturing relationship
\end{itemize}
\vspace{0.3cm}

\noindent \textbf{\texttt{Talks on the Precepts and Buddhist Ethics}}
\begin{itemize}
    \item Conversations centered on Buddhist precepts and ethics.
    \item Discussion on the ethical guidelines and precepts in Buddhism.
    \item Let's discuss the precepts and moral guidelines in Buddhist teachings.
    \item An in-depth exploration of Buddhist precepts and ethics.
    \item Discussing Buddhist precepts and the ethical guidelines that follow.
    \item A detailed discussion about precepts and ethics in Buddhism.
    \item Unpacking the ethical teachings and precepts of Buddhism.
    \item Looking into the ethical teachings and precepts of Buddhism.
    \item Uncovering the precepts and ethical standards of Buddhism.
    \item Understanding the precepts and ethical values of Buddhism.
    \item An examination of Buddhist precepts and the moral code associated with them.
    \item Understanding the principles and ethical teachings of Buddhism.
    \item Exploring the ethical guidelines and teachings of Buddhism.
    \item A talk about the moral teachings and principles upheld by Buddhists.
    \item Let's delve into the teachings and moral code of Buddhism.
    \item Discussing the principles and ethics in Buddhism.
    \item A comprehensive discussion on the principles and ethical standards in Buddhism.
    \item A dialogue about the principles and moral standards in Buddhism.
    \item Exploring the ethical teachings and principles upheld in Buddhism.
    \item A conversation about the ethical principles and precepts followed in Buddhism.
    \item Discussing the ethical values and teachings observed in Buddhism.
    \item Get to know the ethical principles and teachings in Buddhism.
    \item A deep dive into the moral standards and precepts in Buddhism.
    \item Examining the moral guidelines and teachings of Buddhism.
    \item Conversations revolving around the moral standards and principles in Buddhism.
\end{itemize}
\vspace{0.3cm}

\noindent \textbf{\texttt{Sony Won't Release <i>The Interview</i> on VOD}}
\begin{itemize}
    \item Sony opts out of releasing The Interview on VOD.
    \item Sony refrains from releasing The Interview on VOD.
    \item Sony denies the release of The Interview on VOD.
    \item Sony will not launch The Interview on VOD.
    \item Sony decides not to launch The Interview on Video on Demand.
    \item Sony decides not to make The Interview available on VOD.
    \item Sony has decided against the VOD release of The Interview.
    \item Sony rules out releasing The Interview on Video on Demand.
    \item Sony has ruled out the VOD release of The Interview.
    \item Sony is not planning to release The Interview on VOD.
    \item Sony has chosen not to put out The Interview on VOD.
    \item Sony has refused to release The Interview on Video on Demand.
    \item Sony abstains from releasing The Interview on Video on Demand.
    \item Sony has declined to release The Interview on Video on Demand.
    \item Sony won’t be putting The Interview out on VOD.
    \item The Interview will not be available on VOD, says Sony.
    \item The VOD release of The Interview has been canceled by Sony.
    \item Sony is not going to make The Interview available on VOD.
    \item The Interview will not be launched on Video on Demand by Sony.
    \item Sony won’t be releasing The Interview on Video on Demand.
    \item The Interview will not be made available on VOD by Sony.
    \item The Interview won’t be released on VOD by Sony.
    \item Sony will not be sending The Interview to VOD.
    \item Sony isn’t going to release The Interview on Video on Demand.
    \item The Interview's VOD release has been declined by Sony.
\end{itemize}
\vspace{0.3cm}

\noindent \textbf{\texttt{<em>South Park: The Stick of Truth</em> Review (Multi-Platform)}}
\begin{itemize}
    \item A multi-platform review of the game South Park: The Stick of Truth.
    \item A roundup review of South Park: The Stick of Truth across multiple platforms.
    \item Review and analysis of South Park: The Stick of Truth for different platforms.
    \item A detailed review of South Park: The Stick of Truth for various platforms.
    \item A comprehensive review of South Park: The Stick of Truth across various gaming platforms.
    \item Reviewing South Park: The Stick of Truth on its various platforms.
    \item A critique of South Park: The Stick of Truth for multiple gaming platforms.
    \item An overview of South Park: The Stick of Truth for various gaming platforms.
    \item A breakdown review of South Park: The Stick of Truth on different platforms.
    \item A thorough review of South Park: The Stick of Truth on various platforms.
    \item Exploring South Park: The Stick of Truth in a multi-platform context.
    \item Assessment of South Park: The Stick of Truth for multiple gaming platforms.
    \item A thorough examination of South Park: The Stick of Truth on multiple platforms.
    \item Evaluating South Park: The Stick of Truth on multiple gaming platforms.
    \item A multi-platform critique of South Park: The Stick of Truth.
    \item Examining South Park: The Stick of Truth on multiple game platforms.
    \item A comprehensive look at South Park: The Stick of Truth across different platforms.
    \item A deep dive into South Park: The Stick of Truth on multiple platforms.
    \item A game analysis for South Park: The Stick of Truth on multiple platforms.
    \item Analyzing the game South Park: The Stick of Truth for multiple platforms.
    \item Analyzing the multi-platform game South Park: The Stick of Truth.
    \item Delving into South Park: The Stick of Truth on multiple platforms.
    \item Rating South Park: The Stick of Truth across various platforms.
    \item Discussing the merits of South Park: The Stick of Truth on different platforms.
    \item Inspection of South Park: The Stick of Truth across several platforms.
\end{itemize}
\vspace{0.2cm}

\noindent \textbf{\texttt{Insights with Laura Powers}}
\begin{itemize}
    \item Gleaning insights from Laura Powers.
    \item Exploring insights with Laura Powers.
    \item Discovering Laura Powers insights.
    \item Taking in insights from Laura Powers.
    \item Learning from Laura Powers insights.
    \item A deep dive into Laura Powers ideas.
    \item Exploring the mind of Laura Powers.
    \item Unearthing the wisdom of Laura Powers.
    \item Immersing in a conversation with Laura Powers.
    \item Understanding the worldview of Laura Powers.
    \item Indulging in an enlightening talk with Laura Powers.
    \item Understanding Laura Powers perspective.
    \item Getting a glimpse into Laura Powers thought process.
    \item Seeking an intellectual exchange with Laura Powers.
    \item Uncovering Laura Powers thoughts and opinions.
    \item Seeking wisdom from Laura Powers.
    \item Engaging in a knowledge sharing session with Laura Powers.
    \item Delving into Laura Powers expertise.
    \item Discussing various topics with Laura Powers.
    \item Getting to know Laura Powers better.
    \item Venturing into the intellectual space of Laura Powers.
    \item Analysing the thought process of Laura Powers.
    \item Hearing Laura Powers point of view.
    \item Dive into the world of Laura Powers.
    \item Having an intellectual conversation with Laura Powers.
\end{itemize}
\vspace{0.2cm}

\noindent \textbf{\texttt{<i>The Long Dark</i> Gets First Trailer, Steam Early Access}}
\begin{itemize}
    \item The Long Dark has its first trailer out, ready for play on Steam Early Access.
    \item The Long Dark has made its first trailer public, also now available on Steam Early Access.
    \item The Long Dark debuts its first trailer and is now part of Steam Early Access.
    \item The Long Dark introduces its first trailer and is now available on Steam Early Access.
    \item The initial trailer of The Long Dark has been released, now on Steam Early Access.
    \item The Long Dark releases its initial trailer and is now live on Steam Early Access.
    \item The Long Dark releases its first trailer and is now included in Steam Early Access.
    \item The first trailer for The Long Dark has been unveiled, now available on Steam Early Access.
    \item The Long Dark has just revealed its first trailer, also now accessible on Steam Early Access.
    \item The Long Dark has launched its first trailer and it's now on Steam Early Access.
    \item The Long Dark premieres its first trailer and is now available for Steam Early Access.
    \item The Long Dark premieres its first trailer and is now featured in Steam Early Access.
    \item The Long Dark just dropped its first trailer and is now playable on Steam Early Access.
    \item The Long Dark unveils its debut trailer and joins the Steam Early Access platform.
    \item The game The Long Dark released its very first trailer and is now in Steam Early Access.
    \item The first trailer for The Long Dark has been launched, now available through Steam Early Access.
    \item An initial trailer for The Long Dark is out, and it's now accessible on Steam Early Access.
    \item The game The Long Dark has launched its first trailer and is now up for Early Access on Steam.
    \item The Long Dark rolls out its first trailer and is now accessible through Steam Early Access.
    \item The initial trailer for The Long Dark is now out and the game is accessible on Steam Early Access.
    \item Steam Early Access just received The Long Dark and its first trailer has been launched.
    \item The Long Dark has just launched its first trailer and is now available for Early Access on Steam.
    \item The video game The Long Dark has released its first trailer and now is part of Steam Early Access.
    \item Steam Early Access now features The Long Dark with its first trailer being out.
    \item Steam Early Access starts hosting The Long Dark, as its first trailer is launched.
\end{itemize}

\end{document}